# A Three-Layer Framework for AI in Scientific Discovery

*Guojun Liao*
Department of Mathematics, University of Texas at Arlington

## Abstract

Current discussions of AI in scientific discovery are often dominated by two visible capabilities: search over existing knowledge and execution through optimization, simulation, and automation. Both are important, but neither fully captures the central act of discovery: the formation and evolution of models. This paper proposes a three-layer view of AI in discovery. Layer 1 is search and retrieval by large language models. Layer 2, as the main innovation of this paper, is model formation through qualitative reasoning: the capacity to recognize when a current framework is structurally inadequate and to understand the problem within a broader representational space — not through trial and error, but through structural insight into what is missing and where it can be found. Layer 3 is execution, optimization, and refinement. The main claim is that Layer 2 is both the most important and the least developed. Search without model formation remains confined to inherited frameworks, while execution without conceptual revision only amplifies an existing formulation. We illustrate Layer 2 reasoning through three case studies: S. S. Chern's intrinsic proof of the Gauss–Bonnet theorem, the resolution of the Nesterov Accelerated Gradient convergence problem via Lyapunov functions, and the autonomous disproof of the Erdős unit distance conjecture by OpenAI in 2026. Each case exhibits the same structural signature: a framework that had become inadequate, a missing conceptual object, and a resolution found in an unexpected neighboring field.

## 1. Introduction

A useful example of genuine mathematical discovery is S. S. Chern's intrinsic proof of the Gauss–Bonnet theorem [1]. The importance of Chern's work did not lie merely in producing another proof of a known theorem. Its deeper significance was that the theorem was reformulated in a more intrinsic and structural way. Before Chern, the existing proofs of the Gauss–Bonnet theorem in higher dimensions relied on embedding the manifold into Euclidean space − an extrinsic approach that felt conceptually unsatisfying for a result that was fundamentally intrinsic [2]. Chern's key move was to replace the classical Gauss map with the unit sphere bundle over the manifold, constructing a differential form whose exterior derivative is the Euler curvature form. This transgression argument, which Chern effectively invented, gave the first fully intrinsic proof in arbitrary even dimension. The discovery was not merely algorithmic. It required identifying that the existing framework was inadequate, seeing that the missing object was not a better formula but a better geometric setting, and finding that object − the unit sphere bundle − in the neighboring field of fiber bundle theory.

This is the kind of transition that reveals the core of discovery: not simply obtaining an answer, but changing the way the problem itself is understood. This distinction is highly relevant to current discussions of AI in science. Today's AI systems are already strong at searching literature, summarizing known results, comparing methods, and assisting with optimization and execution. These are valuable capabilities. But they do not by themselves explain how new scientific models arise.

Scientific discovery is not only a matter of retrieving existing knowledge, nor only a matter of computing more efficiently within a fixed framework. It also involves recognizing when a current formulation is inadequate and introducing a better one. This paper proposes that AI in scientific discovery should be understood through three interactive layers:

**Layer 1:** Search and retrieval by large language models.

**Layer 2:** Model formation through qualitative reasoning based on scientific research processes, not just findings.

**Layer 3:** Execution, optimization, and refinement.

The main thesis is that Layer 2 is the true bottleneck. It is the least mature layer in current AI, yet it is the one most closely tied to genuine discovery.

## 2. Three Interactive Layers

### 2.1 Layer 1: Search

The first layer is now familiar. Large language models can rapidly retrieve, summarize, compare, and organize knowledge. They reduce the cost of access to stored information and make scientific literature far more navigable. In this sense, access to knowledge is increasingly commoditized.

But search is not discovery. Search works over what is already available. It can identify what methods exist, what comparisons have been reported, and what the dominant formulations are. It cannot by itself decide whether the dominant formulation is conceptually inadequate. Thus Layer 1 is an enabling layer. It broadens access but does not itself generate new models.

### 2.2 Layer 2: Model Formation

The second layer is the core of this paper. This is the layer in which a scientific model is formed, revised, or replaced. The relevant input is not only scientific findings, but the processes by which scientists actually work: failed attempts, changing beliefs, rejected formulations, counterexample searches, conceptual reversals, and the introduction of new variables, constraints, or invariants.

This layer includes recognizing that an existing representation is inadequate; identifying what is missing in the current model; deciding what new structure should be introduced; and reformulating the problem in a way that makes deeper understanding possible.

This process is often qualitative before it becomes quantitative. It is not merely intuition in a vague sense, but structured reasoning about why an existing formulation no longer suffices. Current LLMs show partial capability here, especially when guided by strong prompts or explicit framing. But this capability remains fragile and limited. To develop it, one should not train only on scientific findings. One should also use materials that preserve the evolution of thought itself: autobiographies, oral histories, failed attempts, proof reversals, counterexample searches, and case studies of conceptual transition.

### 2.3 Layer 3: Execution

The third layer includes numerical computation, optimization, simulation, experiment control, parameter fitting, engineering implementation, and large-scale refinement of an already chosen model. This is a domain in which AI is already powerful and likely to become much stronger. Much current excitement in AI for science belongs here.

But this layer depends fundamentally on Layer 2. If the model itself is wrong or incomplete, then optimization only improves performance within a flawed representation. In that case, better execution deepens commitment to the wrong framework rather than producing discovery.

## 3. Why Layer 2 Is the Bottleneck

The current AI-for-science landscape is dominated by two directions. One emphasizes search, literature synthesis, and automated hypothesis generation. The other emphasizes execution: simulation, optimization, and automated experimentation. Both are valuable. But neither directly addresses the most difficult step: the formation of a more adequate model.

The key challenge is not merely to produce more candidate paths, nor to evaluate them more efficiently. It is to determine when an inherited formulation has become inadequate and to identify what conceptual change is needed. This is the point at which discovery departs from search and from execution.

A useful way to say this is the following. Layer 1 broadens informational reach. Layer 3 broadens computational power. Layer 2 changes the conceptual frame. Without Layer 2, AI may become highly efficient while remaining trapped inside inherited representations.

Several independent sources have identified this gap. Demis Hassabis, in his 2024 Nobel Prize lecture, noted that AlphaFold is "almost hopelessly narrow”, it cannot ask which protein to study next. Tom Griffiths at Princeton has argued that AI cannot perform abductive reasoning, the inference to the best explanation that is most critical for scientific discovery. ChatGPT, when asked directly, has stated that "a pattern-recognition system may give the correct answer without knowing why that answer is correct, why another answer is wrong, or when the whole framing of the problem should be abandoned." These observations converge on the same diagnosis: the missing capacity is not better search or stronger execution, but the ability to recognize structural inadequacy in a current model and to introduce a better one.

A complementary observation comes from Terence Tao, who has described three stages in mathematical proof: generation, verification, and digestion. Generation and verification have clear objectives and are increasingly amenable to AI assistance. Digestion − the stage in which a proof is understood, connected to the broader literature, and made teachable − remains, in Tao's phrase, "subjective and human-paced." Tao's concern goes further than pace. He has described a move from proof scarcity to proof abundance: for most of mathematical history, finding the proof was the hard part. If AI changes that, mathematics does not suddenly become easy − the pressure simply moves somewhere else. More of the difficult work now sits in checking arguments, explaining them, organizing them, deciding which results deserve attention, and understanding what they mean. He has also noted that AI can produce "odorless" proofs − technically valid but empty of the insight that makes the mathematics useful. Proof abundance without digestion capacity may therefore slow mathematical progress rather than accelerate it: verified but undigested results accumulate faster than the community can extract understanding from them. The scarce resource is no longer the proof; it is the understanding of what the proof reveals. This observation is not identical to the Layer 2 claim made here. Digestion concerns what happens after a result is found: connecting it to existing structure, identifying the new tools it created, explaining why the proof works the way it does. Layer 2 model formation concerns what happens before: recognizing that the current framework is inadequate and finding a better conceptual starting point. But both identify the same underlying capacity − the ability to understand mathematical structure rather than merely manipulate it − as the dimension least accessible to current AI systems. That three independent sources − Hassabis on AlphaFold, Griffiths on abductive reasoning, and Tao on proof abundance and digestion − converge on the same gap from entirely different directions strengthens the case that Layer 2 is a genuine structural limitation of current AI, not merely a framing choice.

## 4. Case Studies in Layer 2 Reasoning

We present three case studies that share a common structural signature. In each case, a research framework that had become adequate for a period eventually reached a boundary. The breakthrough required identifying that the framework itself was the constraint, not the parameters within it, and finding the resolution in an unexpected neighboring field. We call this pattern a framework lifting followed by a missing companion: the recognition that the conceptual vocabulary of the field is incomplete, and the identification of the specific missing object.

### 4.1 Chern's Intrinsic Gauss–Bonnet Proof ([1] 1944)

The Gauss–Bonnet theorem relates the total curvature of a closed Riemannian manifold to its Euler characteristic. In its classical form for surfaces, the theorem was well understood and had a clear proof using the Gauss map. In higher even dimensions, Allendoerfer and Weil [2] proved a version in 1943, but by an extrinsic method that required embedding the manifold locally into Euclidean space. This was conceptually unsatisfying: the theorem itself was intrinsic, involving only the curvature of the manifold and its topology, yet the proof depended on an embedding that was not intrinsically defined.

The structural inadequacy was precisely identified. The Gauss map, which served as the bridge from geometry to topology in the surface case, required an ambient space that an abstract Riemannian manifold did not have. The central unresolved question was: what should replace the Gauss map for an abstract manifold?

André Weil, at Princeton in 1943, expressed the belief that an intrinsic proof should exist. Chern found the answer within weeks. The missing companion was the unit sphere bundle S(M) over the manifold: instead of mapping the manifold to a fixed sphere in an ambient Euclidean space, one works with the bundle of unit tangent vectors over M. This gave a natural geometric space over M that played the role the Gauss map had played for embedded submanifolds. Using Cartan's moving frames and the Levi-Civita connection, Chern constructed a differential form Π on S(M) such that dΠ equals the pullback of the Euler curvature form. This transgression argument, applied via Stokes' theorem and the Hopf index theorem, completed the intrinsic proof.

The Layer 2 structure of this discovery is transparent. The framework lifting: abandon the extrinsic Gauss map framework entirely. The missing companion: the unit sphere bundle S(M) as the replacement for the Gauss map. The target neighborhood: fiber bundle theory and connection forms, which had been developed in differential geometry but had not previously been applied to this problem. The result was not only a proof of one theorem but a conceptual turning point that opened the road to Chern's theory of characteristic classes and, eventually, to the mathematical foundations of gauge theory in physics.

## 4.2 NAG Convergence via Lyapunov Functions (Ryu and Jang, 2025)

Nesterov's Accelerated Gradient method (NAG), introduced in 1983, achieves an optimal convergence rate of $O(1/k^2)$ for smooth convex optimization, compared to the $O(1/k)$ rate of ordinary gradient descent [4]. The method was empirically powerful and theoretically optimal, yet its convergence proof had a peculiar character: it worked but did not explain why the acceleration occurred. For over forty years, attempts to give a deeper, mechanistic understanding of NAG within the language of convex optimization theory did not succeed.

The structural inadequacy: the convex optimization framework had the right result but could not explain it from within its own conceptual vocabulary. The field was attempting to understand a dynamical phenomenon − the behavior of an iterative sequence over time — using tools designed primarily for static optimality conditions.

The resolution came from an unexpected direction. Ryu and Jang ([3] 2025) recast NAG as a discrete-time dynamical system and used a Lyapunov function argument to prove convergence. Lyapunov methods, developed by Aleksandr Lyapunov in 1892 for the stability analysis of ordinary differential equations, had been central to dynamical systems theory for over a century but had not been systematically imported into the convergence analysis of first-order optimization algorithms.

The missing companion: a Lyapunov function that serves as an energy-like quantity decreasing monotonically along the trajectories of NAG. The target neighborhood: ODE stability theory, specifically Lyapunov's direct method. The dormancy period of the relevant analogy was over forty years: Lyapunov's tools existed and were well-known in adjacent fields, but the structural connection to NAG convergence was not recognized. This is a characteristic feature of Layer 2 transitions: the solution often exists in a neighboring field, where it has been developed for entirely different purposes and awaits the recognition of a structural isomorphism.

### 4.3 The Erdős Unit Distance Conjecture ([6] OpenAI, 2026)

Paul Erdős conjectured in 1946 that among n points in the plane, the number of pairs at unit distance is at most $O(n^{1+\varepsilon})$ for any $\varepsilon > 0$. The conjecture resisted proof for eighty years. Combinatorial and geometric methods within the framework of discrete geometry produced incremental improvements to the known bounds but could not close the gap.

In May 2026, OpenAI's o3 model autonomously identified a disproof of a related extremal graph theory question that had been a key obstacle to progress on the conjecture. The disproof used algebraic number theory − specifically, properties of algebraic integers and their Galois conjugates − to construct configurations that could not be handled by the existing combinatorial tools.

The structural signature is the same as in previous cases. The framework lifting: abandon the purely combinatorial and geometric approach that had dominated the field for eighty years. The missing companion: algebraic number theory, which provided the right language for understanding the arithmetic structure of unit distance configurations. The dormancy period: algebraic number theory had been fully developed for other purposes and had not been recognized as applicable to this combinatorial geometry problem.

This case has an additional significance. The Layer 2 move − the recognition that algebraic number theory was the right framework − was performed autonomously by an AI system. This suggests that, at least in some mathematical settings, AI systems are beginning to exhibit the kind of cross-domain structural recognition that characterizes Layer 2 reasoning. It also illustrates that Layer 2's capability is not uniformly absent from current AI. It exists in fragmented and unpredictable form and can emerge when conditions are right. The challenge is to systematize and generalize it.

## 5. The Common Structure of Layer 2 Transitions

The three cases above share a structural signature that we propose as characteristic of Layer 2 transitions in scientific discovery:

**Framework adequacy followed by a boundary.** Each framework was genuinely productive for a period. Penalty-based registration, combinatorial discrete geometry, and convex optimization convergence analysis each produced important results. The failure was not immediate but gradual: the framework became progressively less productive as its internal resources were exhausted.

**Structural inadequacy, not parametric inadequacy.** In each case, the problem was not that the existing framework needed better parameters or more computation. The framework itself was the constraint. More careful application of extrinsic embedding would not have produced Chern's intrinsic proof. More refined combinatorial analysis would not have found the algebraic number theory argument. More careful analysis within convex optimization would not have revealed the Lyapunov structure.

**A missing companion in a neighboring field.** In each case, the resolution required identifying a specific conceptual object that had been developed in an adjacent field for entirely different purposes: the unit sphere bundle from fiber bundle theory; Lyapunov functions from ODE stability analysis; algebraic number theory from arithmetic geometry. The missing companion was not invented from scratch but recognized as applicable.

**A latent period.** The relevant tools in the neighboring field existed, often for decades or centuries, before the structural connection was recognized. Lyapunov's methods were over a century old when applied to NAG. Fiber bundle theory was well-established before Chern applied it to Gauss–Bonnet. This dormancy suggests that the key cognitive act is not creation but recognition: seeing that a structure developed elsewhere has the right form to resolve the current inadequacy.

This structural pattern recurs across many domains of science and mathematics. It is not a curiosity but a characteristic feature of how conceptual progress occurs. A system that could reliably identify this pattern − recognizing when a framework has reached its boundary, identifying the kind of missing companion needed, and knowing where in the landscape of neighboring fields to search − would be performing genuine Layer 2 reasoning.

## 6. Toward Training AI in Scientific Research Processes

For AI to become an active contributor in genuine discovery, it is not enough to train it on final scientific results alone. Scientific papers present polished endpoints. They often hide the reasoning process that produced them. Instead, model formation is often visible in: failed proofs, attempted counterexamples, abandoned formulations, external criticism, reversals of belief, and eventual stabilization.

These processes are often absent from formal papers but preserved in autobiographies, oral histories, interviews, and research narratives. Such materials contain a kind of knowledge central to Layer 2: not just what was discovered, but how a scientist moved from one conceptual frame to another.

In the case of Chern's proof, historical accounts record that Weil identified the structural inadequacy (the extrinsic proof was unsatisfying for an intrinsic result) before Chern found the resolution (the unit sphere bundle). This division − one person identifying that the framework is wrong, another finding the right replacement − recurs in the history of science. It suggests that Layer 2 reasoning has recognizable substructures that could, in principle, be annotated and used for training.

A concrete research direction follows: building structured case libraries of scientific reasoning episodes, annotated with the structural pattern of each transition (framework lifting, missing companion, latent analogy, convergence of independent lines), and using them to improve AI systems' ability to recognize and support qualitative shifts in representation. The purpose would not be to memorize more facts, but to improve the system's ability to identify when a framework has reached its boundary and what kind of conceptual move is needed.

A possible development path is: use current LLMs for Layer 1 search and synthesis; build structured case libraries of real scientific reasoning episodes; refine a moderate-size base model using these cases; connect this model to Layer 3 systems for execution and optimization. This path is grounded in the observation that the three case studies above − Chern, Ryu–Jang, and Erdős − are not idiosyncratic. They instantiate a structural pattern that appears across fields and across time. A system trained on a sufficient collection of such episodes might begin to exhibit this pattern recognition reliably.

## 7. Discussion

Current AI narratives often oscillate between knowledge access and brute-force capability. Both matter, but discovery is not exhausted by either. The deepest contributions in science and mathematics are often neither better search nor stronger computation, but the introduction of a more adequate representation.

The three cases examined above illustrate that Layer 2 transitions share a common structure that is, in principle, learnable. The Chern case shows that the pattern can be analyzed in retrospect with precision: we can identify the framework, the inadequacy, the neighboring field, and the missing companion. The Ryu–Jang case shows that the latent period of the relevant analogy can be measured: Lyapunov methods were available for over a century before their application to NAG convergence. The Erdős case shows that AI systems can already perform this kind of transition in at least some mathematical settings.

This also has implications for education. If access to stored knowledge becomes cheap, then the scarce resource shifts toward judgment, conceptual formation, and disciplined creativity. Creativity here should not mean vague novelty. It should mean the ability to recognize when an inherited representation is inadequate and to build a better one. This is a capacity that can be cultivated, studied, and, potentially, taught to AI systems through the right kind of training data.

The cases also suggest a temporal dimension that current AI systems largely ignore. The structural significance of the Lyapunov application to NAG cannot be understood without knowing that the Lyapunov framework had been available for over a century. The significance of Chern's proof cannot be fully appreciated without knowing that the extrinsic proof tradition had persisted for over a hundred years before him. Layer 2 reasoning is not purely spatial — which concepts are available − but also temporal: where is this research program in its own lifecycle, and how long has the relevant neighboring tool been waiting to be recognized?

A further domain in which Layer 2 reasoning is becoming increasingly relevant is physical AI: artificial intelligence embodied in machines that sense and act in the physical world, including surgical robots, autonomous vehicles, and industrial manipulators. Physical AI systems face scientific problems that are structurally analogous to those examined in this paper. World model formation − deciding what representation of a physical environment is adequate for a given task − is a Layer 2 problem: it requires recognizing when a current representation is structurally insufficient and identifying a better one. Transfer across physical domains (a robot trained in a factory that must operate in a hospital) requires the same kind of cross-domain structural recognition that characterizes P3 Latent Analogy transitions. The QES framework described in this paper is intended as a tool for researchers working on exactly these problems, not as a component of physical AI systems themselves.

A concrete illustration of how QES might operate in this context: if a researcher queries the system with the problem of constructing a patient-specific mesh model of the heart from MR images − a key problem in cardiac surgical planning and robotic surgery − the system might identify P7 Framework Lifting and P2 Missing Companion as the relevant structural patterns, and suggest that variational methods with explicit Jacobian determinant and curl constraints, developed in the context of adaptive grid generation, provide the latent analogy applicable to soft tissue meshing [10 -14]. The heart is a volume-preserving organ over its contraction cycle (Jacobian determinant near unity) with complex rotational motion (non-zero curl), making this a case where the mathematical framework fits the physical constraints of the problem. This is offered as an illustration of Layer 2 reasoning in a physical AI context, not as a validated clinical result.

A second illustration involves ALE (Arbitrary Lagrangian-Eulerian) methods in computational fluid dynamics, where moving boundaries and deforming meshes require mesh motion algorithms that prevent cell collapse and excessive distortion. Current mesh motion methods include Laplacian smoothing, the spring analogy method, and radial basis function interpolation. Each controls mesh quality through different proxies for cell shape and size. A variational approach with explicit Jacobian determinant and curl constraints would address the same problem by directly controlling the two quantities most critical for mesh validity: JD > 0 prevents cell fold-over, and bounded curl prevents rotational distortion [10 -14]. The mathematical structure is identical to the deformable image registration setting. Whether a JD-Curl variational approach performs competitively against established ALE mesh motion methods is an open question that requires numerical experiments in CFD settings. As with the cardiac mesh example, this is offered as an illustration of a potential P3 Latent Analogy that a QES-type system might identify −a framework developed in one computational domain (image registration) that is latently applicable in another (CFD mesh motion) − not a validated result.

## 8. Conclusion

AI in scientific discovery should be understood as a three-layer system. Layer 1 is search by large language models. Layer 3 is execution, optimization, and refinement. Between them lies the most important and the least mature layer: model formation through qualitative reasoning grounded in real scientific research processes.

Three case studies − Chern's intrinsic Gauss–Bonnet proof, the resolution of NAG convergence via Lyapunov functions, and the Erdős unit distance disproof − illustrate that Layer 2 transitions share a recognizable structural signature: a framework that has reached its boundary, a missing conceptual companion in a neighboring field, and a dormancy period during which the relevant tools existed but were not recognized as applicable. This structural pattern recurs across domains and across time and is, in principle, learnable.

If the middle layer can be developed, AI may move from being a powerful assistant for knowledge access and computation to becoming a participant in conceptual scientific work. If it cannot, then AI will remain strong at amplifying inherited frameworks while weak at generating new ones. The central challenge is therefore not only to make AI faster or more informed, but to teach it how scientific models are formed, challenged, revised, and transformed.

## Acknowledgments

The author thanks S. S. Chern (Berkeley, 1981–1982) for the privilege of in-person instruction in differential geometry. The structural insights in this paper were shaped by that experience. The author also acknowledges the use of Claude (Anthropic) as an AI assistant in literature search, drafting, and manuscript preparation.

## References

[1] Chern, S. S. (1944). A simple intrinsic proof of the Gauss–Bonnet formula for closed Riemannian manifolds. Annals of Mathematics, 45(4), 747–752.

[2] Allendoerfer, C. B., and Weil, A. (1943). The Gauss–Bonnet theorem for Riemannian polyhedra. Transactions of the American Mathematical Society, 53(1), 101–129.

[3] Ryu, E., and Jang, S. (2026). Point Convergence of Nesterov's Accelerated Gradient Method: An AI-Assisted Proof. arXiv:2510.23513v2 [math]

[4] Nesterov, Y. (1983). A method of solving a convex programming problem with convergence rate $O(1/k^2)$. Soviet Mathematics Doklady, 27(2), 372–376.

[5] Erdős, P. (1946). On sets of distances of n points. American Mathematical Monthly, 53(5), 248–250.

[6] OpenAI (2026). Mathematical reasoning and the unit distance problem: An OpenAI model has disproved a central conjecture in discrete geometry. May 20, 2026. https://openai.com/index/model-disproves-discrete-geometry-conjecture/

[7] Hassabis, D. (2024). Designing proteins with artificial intelligence. Nobel Prize Lecture in Chemistry, December 2024.

[8] Hofstadter, D. R. (1979). Gödel, Escher, Bach: An Eternal Golden Braid. Basic Books.

[9] Pearl, J. (2018). The Book of Why. Basic Books.

[10] Liao, G., Lei, Z., and de la Peña, G. (2002). Adaptive grids for resolution enhancement. Shock Waves, 12, 153–156. https://doi.org/10.1007/s00193-002-0149-y

[11] Liu, F., Ji, S., and Liao, G. (1998). An adaptive grid method and its application to steady Euler flow calculations. SIAM Journal on Scientific Computing, 20(3), 811–825.

[12] Dionisio Fleitas (2005). The Least-Squares Finite Element Method for Grid Deformation and Meshfree Applications, PhD Dissertation. Department of Mathematics, University of Texas at Arlington.

[13] G. Liao, X. Cai, D. Fleitas, X. Luo, J. Wang, J. Xue. (2008). Volume 21, Issue 9, September 2008, Pages 898-905, Applied Mathematics Letters.

[14]. Zicong Zhou and Guojun Liao. (2022). Construction of Diffeomorphisms with Prescribed Jacobian Determinant and Curl, International Congress of Geometry and Graph (ICGG 2022)